\documentclass[conference]{IEEEtran}
\IEEEoverridecommandlockouts
\usepackage{cite}
\usepackage{amsmath,amssymb,amsfonts}
\usepackage{algorithmic}
\usepackage{booktabs}
\usepackage{graphicx}
\usepackage{hyperref} 
\usepackage{multirow} 
\newtheorem{remark}{Remark}
\usepackage{textcomp}
\usepackage{subfigure}
\usepackage{xcolor}
\def\BibTeX{{\rm B\kern-.05em{\sc i\kern-.025em b}\kern-.08em
    T\kern-.1667em\lower.7ex\hbox{E}\kern-.125emX}}
\begin{document}

\title{TYrPPG: Uncomplicated and Enhanced Learning Capability rPPG for Remote Heart Rate Estimation
}

\author{\IEEEauthorblockN{Taixi Chen$^{1,}$$^{2}$  \qquad
Yiu-ming Cheung$^{2}$\IEEEauthorrefmark{2}\thanks{\IEEEauthorrefmark{2}Corresponding author}\thanks{This work is supported by RGC Senior Research Fellow Scheme under grant: SRFS2324-2S02.}}
\IEEEauthorblockA{$^{1}$School of Computing,
Binghamton University, Binghamton, NY, USA}
\IEEEauthorblockA{$^{2}$Department of Computer Science, Hong Kong Baptist University, Hong Kong SAR, China}
\IEEEauthorblockA{tchen51@binghamton.edu; ymc@comp.hkbu.edu.hk}}

\maketitle
\begin{abstract}
Remote photoplethysmography (rPPG) can remotely extract physiological signals from RGB video, which has many advantages in detecting heart rate, such as low cost and no invasion to patients. The existing rPPG model is usually based on the transformer module, which has low computation efficiency. Recently, the Mamba model has garnered increasing attention due to its efficient performance in natural language processing tasks, demonstrating potential as a substitute for transformer-based algorithms. However, the Mambaout model and its variants prove that the SSM module, which is the core component of the Mamba model, is unnecessary for the vision task. Therefore, we hope to prove the feasibility of using the Mambaout-based module to remotely learn the heart rate. Specifically, we propose a novel rPPG algorithm called uncomplicated and enhanced learning capability rPPG (TYrPPG). This paper introduces an innovative gated video understanding block (GVB) designed for efficient analysis of RGB videos. Based on the Mambaout structure, this block integrates 2D-CNN and 3D-CNN to enhance video understanding for analysis. In addition, we propose a comprehensive supervised loss function (CSL) to improve the model’s learning capability, along with its weakly supervised variants. The experiments show that our TYrPPG can achieve state-of-the-art performance in commonly used datasets, indicating its prospects and superiority in remote heart rate estimation. The source code is available at \href{https://github.com/Taixi-CHEN/TYrPPG}{https://github.com/Taixi-CHEN/TYrPPG.}
\end{abstract}

\section{Introduction}

\begin{figure*}[t]
\centering
\subfigure[TYrPPG-KL]{
    \includegraphics[height = 33mm]{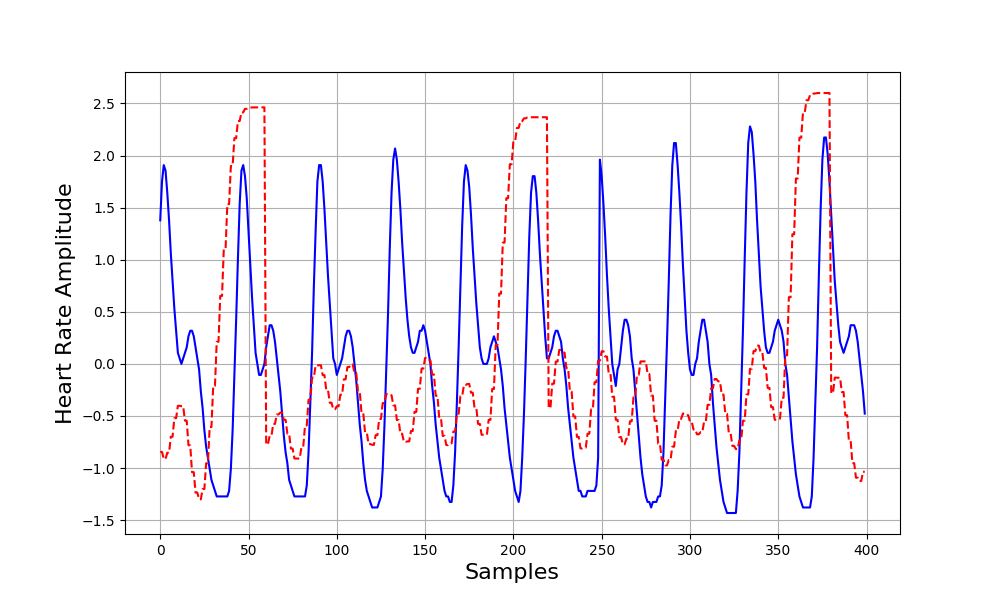}
}
\subfigure[TYrPPG-video MMD]{
    \includegraphics[height = 33mm]{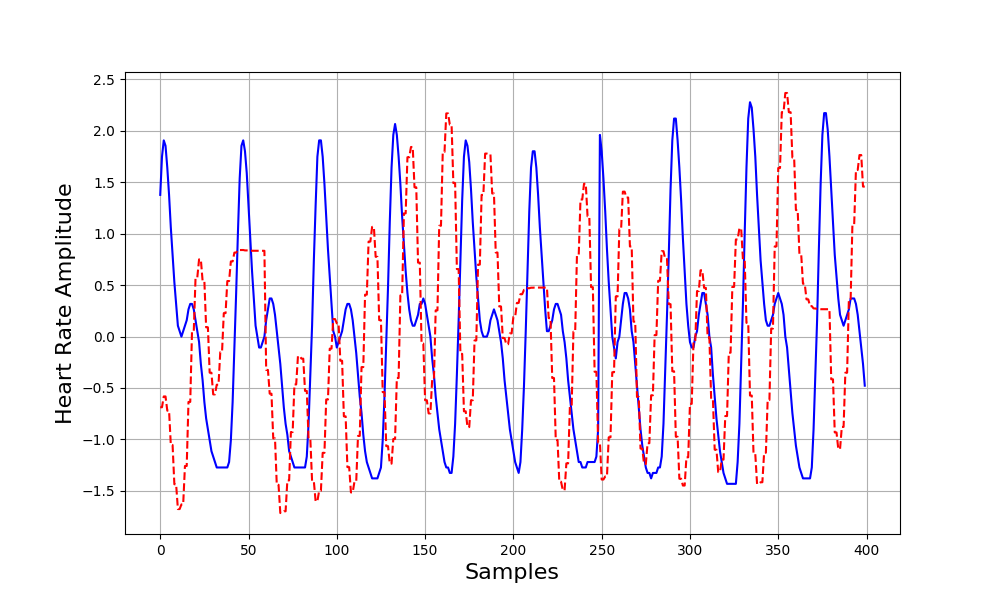}
}

    \caption{Visualization of the heart rate signals estimated by TYrPPG based on KL divergence and proposed video MMD. Straight and dotted lines mark the signals as ground truth and model estimation, respectively. TYrPPG can learn the ground truth better optimized by video MMD, as its estimation signal peaks are more consistent with the ground truth. This shows the effectiveness of the proposed video MMD.}
    \label{subfig:klmmd}
\end{figure*}

Remote heart rate detection has been extensively studied due to its ability to eliminate concerns related to physical contact between doctors and patients \cite{Ba, Ver}. This aspect has become increasingly important in the post-COVID-19 era. Besides, it can be leveraged in affective computing and deepfake detection \cite{rhythmmamba}, showing its wide usage domain. In particular, remote photoplethysmography (rPPG) is a typical method for remotely estimating the heart rate. It extracts the Blood Volume Pulse (BVP) to isolate the heart rate by capturing facial color changes resulting from periodic blood circulation. Traditional rPPG algorithms have made valuable contributions to the healthcare aspect. However, most conventional research and tests are conducted in optimal environments, leading to low accuracy and a lack of robust extraction methods under complex conditions. To address these limitations, both unsupervised and supervised learning approaches have been proposed to mitigate the adverse effects of the environment and head motion.

In unsupervised approaches, the absence of well-rounded pre-assumptions significantly restricts their performance. \cite{motion}. Moreover, some unsupervised domain adaptation methods \cite{Du, toalign} aim to increase model generalization ability but may create distorted images without the target domain label. The supervised approaches are commonly based on deep learning, which needs a large amount of video data for training and myriad times for model convergence as well. Thus, they suffer expensive costs in video analysis and denoise \cite{andrppg}. These complicated structures restrict the possibility of deploying the model in portable devices for widespread use. DeepPhys \cite{solve} was proposed to leverage the normalized light difference to alleviate the negative impacts of various illumination cases. Besides, numerous experiments proved that the attention mechanism can address head motion issues \cite{962}. 

Therefore, existing methods have at least two drawbacks as follows: 1) Using a complicated module in this medical task, which makes them difficult to deploy in a portable device, and 2) Optimizing their model solely based on the unsupervised loss, such as contrastive learning, neglecting to leverage comprehensive information from ground truth. Recently, the Mamba model~\cite{mamba} has demonstrated its better capabilities in natural language tasks and has also proven effective in vision tasks. However, Mambaout \cite{mambaout} has shown that its core component, the SSM module, can be removed for vision tasks, reducing computational and resource costs. 

\begin{figure*}[t]
\centering
\includegraphics[width=.75\linewidth]{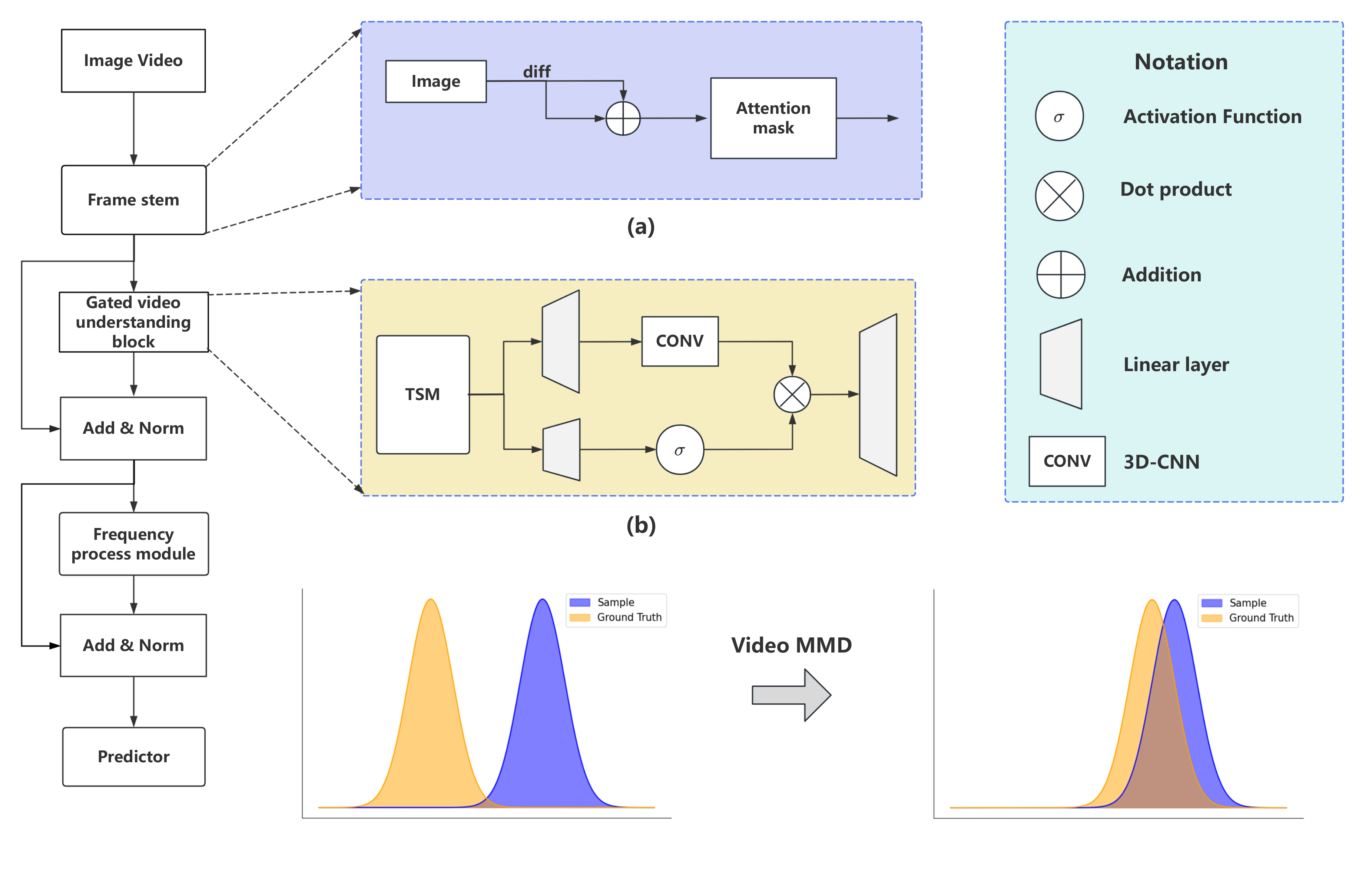}      
\caption{TYrPPG is a gated 3D-CNN-based model structure. (a) shows the frame stem that is a data augmentation block to help TYrPPG understand the video better. (b) is the GVB, containing a TSM module and a gated 3D-CNN, which is designed to efficiently analyze video. Lastly, the distribution dissimilarity learning process is based on our proposed video MMD. Thus, the whole model structure is simpler than the Mamba-based and transformer-based models.}
    \label{fig:model}
\end{figure*}

Thus, our paper aims to propose an effective method called: Uncomplica\textbf{T}ed and enhanced learning capabilit\textbf{Y} \textbf{rPPG} algorithm (\textbf{TYrPPG}) that solves the remote heart rate estimation task, based on Mambaout structure. Due to the rPPG signal being periodic, rough physiological signals are not difficult to extract in most cases. Thus, the motivation of TYrPPG is to prove whether the model can estimate the heart rate with a simpler network structure and discover a path to enhance the model's learning ability. Specifically, due to the success of the TSM \cite{TSM} and Mambaout structure \cite{mambaout}, we create a novel gated video understanding block (GVB) based on them. With the GVB, our model can more effectively analyze facial videos and estimate heart rate using a simpler neural network. To the best of our knowledge, this is the first rPPG algorithm utilizing the Mambaout-based structure, distinguishing it from the recently popular transformer-based and Mamba-based models. Figure~\ref{fig:model} shows the model structure.

Moreover, the TYrPPG framework proposes a novel Comprehensive Supervised Loss (CSL). We hope to improve the ability to learn ground truth distributions by using the proposed CSL. Unlike existing algorithms that utilize KL divergence, we are the first to introduce Maximum Mean Discrepancy (MMD) \cite{MMD} as a component of the proposed Loss function. To avoid over-fitting and remove temporal redundancy, we propose a Video-MMD, which makes the model better learn the discrepancy between their estimation signals and ground truth from video, inspired by \cite{physformer}. Due to the complexity of the rPPG signals, models often predict many incorrect values. When using KL divergence, these discrepancies can be excessively exaggerated, leading to infinite results, which is not desirable. In contrast, MMD can mitigate this issue effectively. Figure~\ref{subfig:klmmd} also proves that its performance surpasses the commonly used KL divergence. In addition, we construct a Weak Supervised Loss (WSL) to utilize only limited distribution information for optimization. This logic is shown in Figure~\ref{fig:logic}. Therefore, this paper makes contributions as below:
\begin{itemize}
\item Proposing an efficient gated video understanding block (GVB) to robustly understand the rPPG video and estimate the rPPG signal, which comprises the TSM and modified Mambaout module. It is a combination of 2D-CNN and 3D-CNN without complex structures.

\item Based on the proposed video-MMD, designing a novel Comprehensive Supervised Loss (CSL) and its variant Weak Supervised Loss (WSL) to utilize the important information from ground truth distribution, leading to fast and accurate convergence.

\item Experiments show TYrPPG can successfully remotely estimate heart rate and obtain a good generalization ability, achieving state-of-the-art performance in the PURE and MMPD datasets, which proves the superiority and promising future of our model.

\end{itemize}

\section{Related Works}
\label{related work}
The conventional rPPG algorithms have proved the feasibility of remotely detecting heart rate using a web camera through RGB channels \cite{Ba, Ver}. It is also known that the green channel should be more suitable for estimating the rPPG signal \cite{Ver}. 

\subsection{Supervised rPPG Algorithm}
Conversely, the deep learning-based methods show their power in complicated conditions and domain generalization ability. As the previous section mentioned, the unsupervised domain adaptation method has a stronger generalization ability but may create a distorted image \cite{Du}. The deep learning-based algorithm \cite{solve} can rely on the ground truth to learn more information. They also utilize the normalized light difference as an input, which is defined as:
\begin{equation}
 X_t = \frac{x_{t+1}-x_{t}}{x_{t+1}+x_{t}}.
\label{solve}
\end{equation}
Moreover, to extract the rPPG signal, some researchers proposed the transformer-based \cite{physformer} and mamba-based models \cite{rhythmmamba} to analyze the human facial model, but they ignore the fact that the rPPG signal is periodic and the rough physiological signals are not hard to capture in most cases. Thus, their model costs a bit more and faces over-fitting issues in limited training data conditions and an unseen environment.
\subsection{Mamba Model}
Mamba \cite{mamba} is a new method designed for NLP originally. The core component of Mamba is the State Space Model (SSM), which defines a sequence-to-sequence transformation as:
\begin{equation}
\begin{aligned}
    h'(t) &= Ah(t) + Bx(t), \\
    y(t) &= Ch(t),
\end{aligned}
\end{equation}
where $A, B, C \in \mathbb{R}^{n\times n}$ and they are learnable parameters. Further, Mamba \cite{mamba} uses the gated structure to capture the long-term relationship:
\begin{equation}
    \begin{aligned}
        g_t &= \sigma(Linear(x_t)), \\
    h_t &= (1-g_t)h_{t-1} + g_tx_t.
    \end{aligned}
\end{equation}
Besides, the biggest difference between  Mambaout \cite{mambaout} and Mamba \cite{mamba} is that Mambaout does not use SSM. The other details are illustrated in Section \ref{pm}. Moreover, the proposed method belongs to the supervised approach, and we will focus on this approach.
\section{Proposed Methodology}
\label{pm}
Reviewing the existing methods and their drawbacks, we propose a novel supervised TYrPPG model.
\begin{figure}[t]
    \centering
    \includegraphics[width=0.45\textwidth]{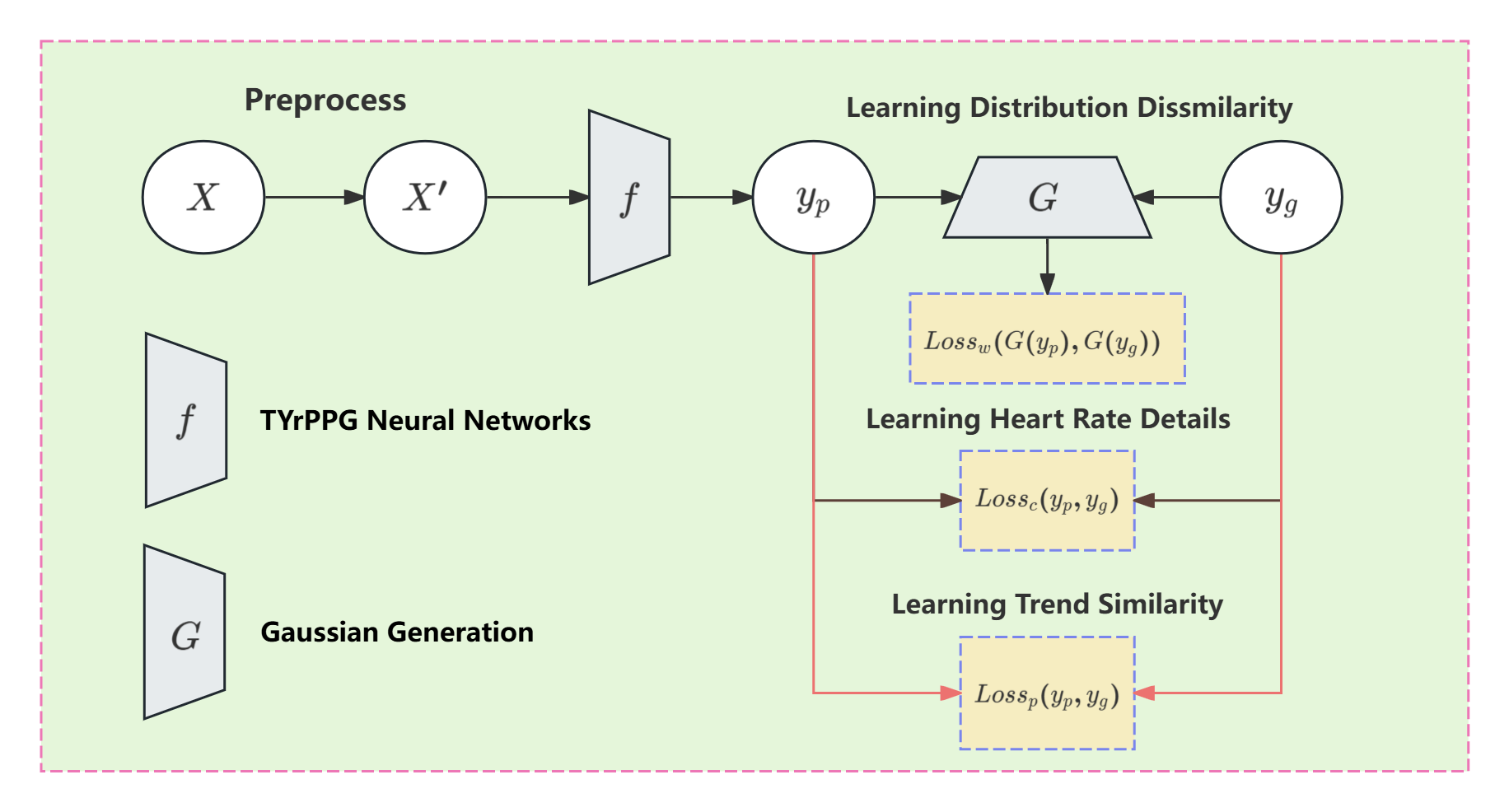}       
    \caption{TYrPPG uses the Comprehensive Supervised Loss (CSL) to optimize the model. CSL contains three loss terms for different purposes, including 1) learning distribution dissimilarity, 2) learning heart rate details, and 3) learning trend similarity. Besides, its variant, Weak Supervised Loss (WSL), comprises only the $\textbf{Loss}_{\textbf{w}}\textbf{(.)}$ and $\textbf{Loss}_{\textbf{p}}\textbf{(.)}$, exploring the feasibility of obtaining 
 a good generalization ability without learning the signal details.}
    \label{fig:logic}
\end{figure}

\subsection{Model strusture}
\begin{table*}[t]
\centering
\caption{Intra-dataset results on PURE and MMPD datasets}
    \begin{tabular}{ c | c c c c c c}
    \toprule
      \multirow{2}{*}{Methods}  & \multicolumn{3}{c}{PURE} &  \multicolumn{3}{c}{MMPD} \\
     \cline{2-7}
       & MAE$\downarrow$ & RMSE$\downarrow$ & $\rho\uparrow$ & MAE$\downarrow$ & RMSE$\downarrow$ & $\rho\uparrow$ \\
        \midrule
         PhysNet \cite{physnet} &  8.21 & 11.35 & 0.19 & 14.46 & 19.45 & 0.19\\
        
        TS-CAN \cite{TS-CAN} & 10.47 & 14.68 & 0.23 & 17.65 & 22.36 & 0.07  \\

        DeepPhys \cite{solve}  & 7.88 & 13.32 & 0.37 & 16.10 & 20.71 & 0.15 \\
        
       PhysFormer  \cite{physformer}  &  5.29 & 6.95 & \underline{0.91} & 15.15 & 19.47 & 0.12  \\

       RhythmFormer \cite{rformer} & \underline{0.97}& \underline{1.10} & \textbf{0.99} & 12.91 & 16.58 & \underline{0.31}\\
       
       \midrule
    \textbf{Ours} & \textbf{0.86} & \textbf{1.02} & \textbf{0.99}& \textbf{12.68}& \textbf{15.71} & \textbf{0.34}  \\
    \bottomrule
    \end{tabular}
  \label{tab:in1}
\end{table*}
\begin{table}[t]
\centering
\caption{Intra-dataset results on the PURE dataset with a train-test split ratio of 6:4.}
    \begin{tabular}{c | c c c}
    \toprule
      Methods & MAE$\downarrow$ & RMSE$\downarrow$ & $\rho\uparrow$ \\
        \midrule
        PhysNet \cite{physnet} &  10.31 & 14.11 & 0.21\\
        
        TS-CAN \cite{TS-CAN} & 11.48 & 15.28 & 0.23  \\

        DeepPhys \cite{solve}  & 10.02 & 13.32 & \underline{0.37}  \\
        
       PhysFormer \cite{physformer}  &  9.67 & \underline{12.68} & 0.21  \\

       RhythmFormer \cite{rformer} & \textbf{8.38}& 13.26 & 0.35 \\
       
       \midrule
    \textbf{Ours} & \underline{9.25} & \textbf{11.51} & \textbf{0.41 }  \\
    \bottomrule
    \end{tabular}
  \label{tab:pure0.6}
\end{table}
The proposed TYrPPG model utilizes the normalized light difference as an input to mitigate the illumination impact by using Eq.(\ref{solve}). Inspired by \cite{solve}, we also utilize the attention mask to capture the face, avoiding the head motion impact, which is computed by:
\begin{equation}
 Mask = \frac{(H/8)(W/8)\cdot\sigma(X)}{2||\sigma(X)||_1},
\label{mask}
\end{equation}
where the $H$ is the height and $W$ is the width of the video frame. $\sigma(.)$ is the activation function. Then, the Gated Video Understanding Blocks (GVB) analyze the video frames to understand the video context for heart rate estimation. It comprises two parts: a TSM module \cite{TSM} based on 2D-CNN and a gated 3D-CNN based on the Mambaout structure \cite{mambaout}. The TSM can first process the video by shifting the channel features in the temporal domain. Assuming that there are T frames, we can get the channel feature vectors of those frames as $X$ = $[X_1, X_2, ..., X_T]$. And the weights of convolution are $W = (w1, w2, w3)$. The motivation for using TSM is that shifting channel features along the time dimension can improve the understanding of video context and make the TYrPPG more robust. The operator is defined as:
\begin{equation}
 Y = w_1X_{t-1} + w_2X_t + w_3X_{t+1},
\label{TSM}
\end{equation}
where the $X_{t-1}$ means the previous frame of the $X_t$, and $X_{t+1}$ is the latter frame. Due to the success of the Mambaout model \cite{mambaout} in vision tasks, we propose a gated 3D-CNN that can be regarded as a variant of it, shown in the model structure in Figure \ref{fig:model} (b). Instead of using the 2D-CNN like the original Mambaout structure, we use a 3D-CNN to replace it, which considers the information from both spatial and temporal domains. Thus, it is more powerful and suitable for analyzing video data compared with the original Mambaout model. The definition of gated 3D-CNN is shown in Eq.(\ref{gate1}) and Eq.(\ref{GVB}).
\begin{table*}[t]
\centering
\caption{Cross-dataset results on PURE and MMPD datasets}
    \begin{tabular}{c | c | c c c c c c}
    \toprule
     \multirow{2}{*}{Methods}&\multirow{2}{*}{TrainSet} & \multicolumn{3}{c}{PURE} &  \multicolumn{3}{c}{MMPD} \\
    \cline{3-8}
      &  & MAE$\downarrow$ & RMSE$\downarrow$ & $\rho\uparrow$ & MAE$\downarrow$ & RMSE$\downarrow$ & $\rho\uparrow$ \\
        \midrule
        \multirow{2}{*}{PhysNet~\cite{physnet}} & PURE  &  - & - & - & 16.43 & 20.87 & 0.01 \\
        & MMPD &  \underline{11.27} & 16.67 & 0.39 & - & - & -\\
        
        \multirow{2}{*}{TS-CAN~\cite{TS-CAN}}  & PURE & - & - & - & 15.65 & 20.86 & 0.13 \\
        &MMPD& 11.31 & 15.41 & 0.13 & - & - & -\\

        \multirow{2}{*}{DeepPhys~\cite{solve}}  & PURE & - & - & - & 15.64 & 20.39 & 0.17 \\
        &MMPD & 12.21 & 16.11 & 0.11 & - & - & - \\
        
        \multirow{2}{*}{PhysFormer~\cite{solve}} & PURE & - & - & - & \underline{15.27} & 19.05 & 0.06  \\
        &MMPD & 11.64 &\textbf{13.56} & \underline{0.25} & - & - & -  \\

        \multirow{2}{*}{RhythmFormer~\cite{rformer} } & PURE & - & - & - & 15.42& \underline{18.79} & \underline{0.24}\\
        & MMPD  & \textbf{10.93} & 15.31 & \underline{0.25} & - & - & - \\
       
       \midrule
     \multirow{2}{*}{\textbf{Ours}}& PURE & - & - & - & \textbf{13.42}& \textbf{16.28} & \textbf{0.32}  \\
     & MMPD & 11.31 & \underline{15.26} & \textbf{0.31}& -& - & - \\
    \bottomrule
    \end{tabular}
  \label{tab:inter1}
\end{table*}
Based on the Mambaout \cite{mambaout}, TYrPPG normalizes input:
\begin{equation}
x' = Norm(x),
\label{normit}
\end{equation}
where $Norm(.)$ is the layer normalization.  Moreover, TYrPPG will split the $x_t$ into three parts: $x_{c}$, $x_i$ and $x_{g}$, which is defined as:
\begin{equation}
\begin{aligned}
   x_g = Linear(x'), \quad x_i = Linear(x'), \quad x_c = Linear(x'),\\
\end{aligned}
\label{g1}
\end{equation}
where $x_g \in D^H$, $x_i \in D^{H-C}$, and $x_c = D^{C}$. This means that the dimension of the $x_{g}$ equals the hidden dimension. $x_{i}$'s dimension equals hidden dimensions subtractive channel dimension, and $x_c$'s dimension equals channel dimension, which will be processed by the 3D-CNN. Then we can constructs $x_{t}'$ as:
\begin{equation}
x_{c}' = Conv(x_c),
\label{gatei}
\end{equation}
where the $Conv(.)$ is the 3D-CNN. Further, we concatenate the $x_i$ and $x_c$:
\begin{equation}
    x_o = Concatenate(x_i, x_c).
\end{equation}
Then, the output processed value is computed by:
\begin{equation}
x_{o1} = W^T_{o1}(\sigma (x_g)*x_o),
\label{gate1}
\end{equation}
Lastly, we utilize the commonly used residual structure to help the model understand the video information better:
\begin{equation}
x_f = x_{o1} + x,
\label{GVB}
\end{equation}
where the $\sigma(.)$ is the activation function and $x_f$ is the final output of the GVB.  

\begin{remark}
The proposed model structure can easily address the head motion issue and illumination impacts without a complex network structure and assumptions.
\end{remark}

\subsection{Loss Function}
Another critical part of the proposed model is our designed loss function. Specifically, the loss function comprises three different types that respectively focus on the signal's details and distributions. The whole training logic based on different loss functions is shown in Figure~\ref{fig:logic}. Specifically, the training method of our TYrPPG model is called Comprehensive Supervised Loss (CSL). We utilize the ground truth in the frequency domain to optimize our model. The first component is computed by: \begin{equation}
    Loss_c = CE(y_{pred}, y_{pos}),
\label{loss_c}   
\end{equation} 
where the $Loss_c$ is a cross-entropy loss to help TYrPPG learn the details of the heart rate signals. Moreover, we also use the negative Pearson correlation to optimize our model to maximize the trend similarity and minimize peak location errors, which is defined as:
\begin{equation}
\begin{aligned}
&P = \frac{T\sum_1^Txy-\sum_1^Tx\sum_1^Ty}{\sqrt{(T\sum_1^Tx^2-(\sum_1^Tx)^2)(T\sum_1^Ty^2-(\sum_1^Ty)^2)}},\\
&Loss_p = 1 - P,
\end{aligned}
\label{person}
\end{equation}
where $p$ is the Pearson correlation. Moreover, we design a weakly supervised loss, video-MMD loss function. The original MMD is computed by:
\begin{equation}
    \begin{aligned}
        MMD[f, p, q] &= sup(E_p[f(x)]-E_q[f(x)])\\
        &=sup<\mu_p-\mu_q, f>_H\\
        &\leq ||\mu_p-\mu_q||_H.
    \end{aligned}
\end{equation}
However, the maximum value of the Power Spectral Density (PSD) is more meaningful for estimating the heart rate. Inspired by \cite{physformer}, to better generalize the distribution of the ground truth, we decide to generate a series of Gaussian distributions based on the maximum values of the PSD, then utilize MMD to optimize our model, which is defined as:
\begin{equation}
\begin{aligned}
Loss_{w} =  MMD(&maxG(PSD(PPG_{gt})),\\
& maxG(PSD(PPG_{pred}))),
\label{weak}
\end{aligned}
\end{equation}
where $maxG(.)$ regards the maximum Power Spectral Density (PSD) value as the mean value to generate the estimated Gaussian distributions and target Gaussian distributions \cite{physformer}. Then, we use the MMD between two generated Gaussian distributions as a weak supervision. In addition, we also construct the WSL that will not consider the $Loss_c$. The motivation is to prove that less supervision may also work in heart rate estimation.
\begin{remark}
    It is noteworthy that the rough physiological signals are not hard to extract in most cases, but the model will inevitably predict some incorrect values that do not exist in the ground truth. In this case, however, KL divergence can excessively exaggerate the difference between the predicted signals and ground truth due to those events with zero probability, causing incorrect optimization direction. Thus, MMD can provide a more robust evaluation for distribution dissimilarity, better learning the peaks and trends.
\end{remark}
In this case, the proposed CSL and WSL are defined as:
\begin{equation}
Loss_{CSL} = \alpha Loss_{c} + \beta Loss_{p} + \gamma Loss_w,
\label{wholeloss1}
\end{equation}
and
\begin{equation}
Loss_{WSL} = \beta Loss_{p} + \gamma Loss_w,
\label{wholeloss2}
\end{equation}
where the $\alpha$, $\beta$, and $\gamma$ are three hyper-parameters. In our experiments, they are set as 1.0, 1.0, and 2.0. 

\section{Experiments}
In the experiment section, we first introduce the experiment settings and then show the experiment results.
\subsection{Experiment Setting}
We use two public datasets to evaluate the performance of TYrPPG: PURE \cite{PURE} and MMPD \cite{MMPD}.

\textbf{PURE} has 60 videos, including 10 different participants and 6 different test conditions (speaking, moving, etc.), with a video resolution of 640x480 and a frame rate of 30Hz.

\textbf{MMPD} including MMPD (370G, 320 x 240 resolution) and Mini-MMPD (48G, 80 x 60 resolution). We chose Mini-MMPD to train our model, which contains 33 subjects. Moreover, this dataset considers more conditions, including skin tone, activities, and lighting.

Besides, we utilize three commonly used indicators to evaluate model performance: Mean Absolute Error (MAE), Root Mean Squared Error (RMSE), and Pearson correlation coefficient ($\rho$). We set the number of epochs to 30 and the learning rate to $1\times10^{-4}$. We choose the best results of the models to report. Further, the bold means the best, and the underline means the second best.
\subsection{Comparative results}
We conduct two main experiments to evaluate TYrPPG, including the intra-dataset and cross-dataset experiments. In particular, we choose some baseline models to make the comparison:
PhysNet \cite{physnet}, DeepPhys \cite{solve}, physFormer \cite{physformer}, TS-CAN \cite{TS-CAN}, and RhythmFormer \cite{rformer}. 

In the first intra-dataset experiment, we split the PURE dataset and MMPD dataset for training and testing. In particular, the first 80\% of the PURE dataset is regarded as the training data, and the last 20\% of the PURE dataset was utilized for testing. Besides, the ratio of splitting the MMPD dataset is 1:1. Based on Table \ref{tab:in1}, it is clear that our model reaches state-of-the-art performance in the PURE and MMPD datasets. TYrPPG outperforms all supervised methods. It especially outperforms the transformer-based SOTA models, such as Physformer \cite{physformer} and RhythmFormer \cite{rformer}, which proves that the Mambaout-based GVB structure is more powerful compared with the transformer-based model. Moreover, in the second intra-dataset experiment, we also conduct another experiment on the PURE datasets and set its train-test split ratio to 6:4 according to \cite{physformer} to further evaluate the performance of the TYrPPG. Based on Table \ref{tab:pure0.6}, we discover that the advantage of TYrPPG is also obvious, as it ranks first in RMSE and Pearson indicator, and ranks second in MAE indicator.

\begin{table}[t]
\centering
 \caption{Ablation study results on PURE, whose train-test split ratio is 6:4. The results reflect the effectiveness of the proposed CSL and prove the prospect of the WSL.}
    \begin{tabular}{c | c c c}
    \toprule
      Loss terms & MAE $\downarrow$ & RMSE $\downarrow$ & $\rho \uparrow$ \\
        \midrule
        $Loss_c$ & 13.41 & 18.58 & 0.16 \\
        $Loss_t$ & \underline{11.60} & 16.54 & 0.07 \\
        $Loss_h$ & 18.69 & 22.03 & 0.04 \\
        $Loss_h + Loss_c$ & 13.40 & 18.58 & \underline{0.19} \\
        $Loss_c + Loss_t$ &  12.16&  15.84&  0.16\\
        \midrule
        $Loss_h + Loss_t$ (WSL) & \underline{11.60} & \underline{15.57} & 0.04 \\
        $Loss_c + Loss_t + Loss_h$ (CSL) & $\textbf{9.25}$ & \textbf{11.51} & \textbf{0.41} \\
    \bottomrule
    \end{tabular}
  \label{tab:ablation}
\end{table}
\begin{figure*}[t]
\centering
\subfigure[Physformer]{
    \includegraphics[height = 35mm, width = 55mm]{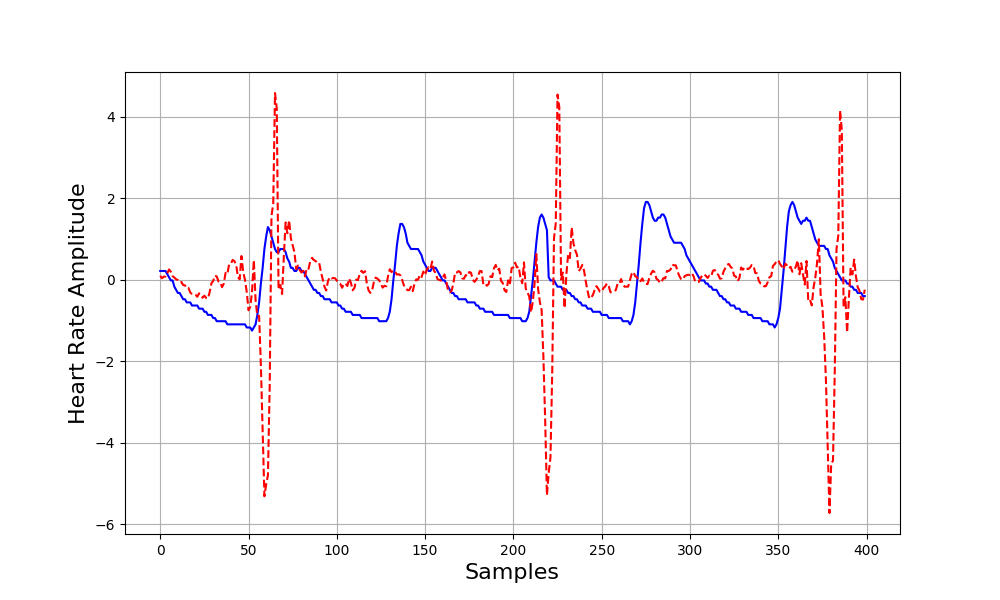}
}
\subfigure[RythmFormer]{
    \includegraphics[height = 35mm, width = 55mm]{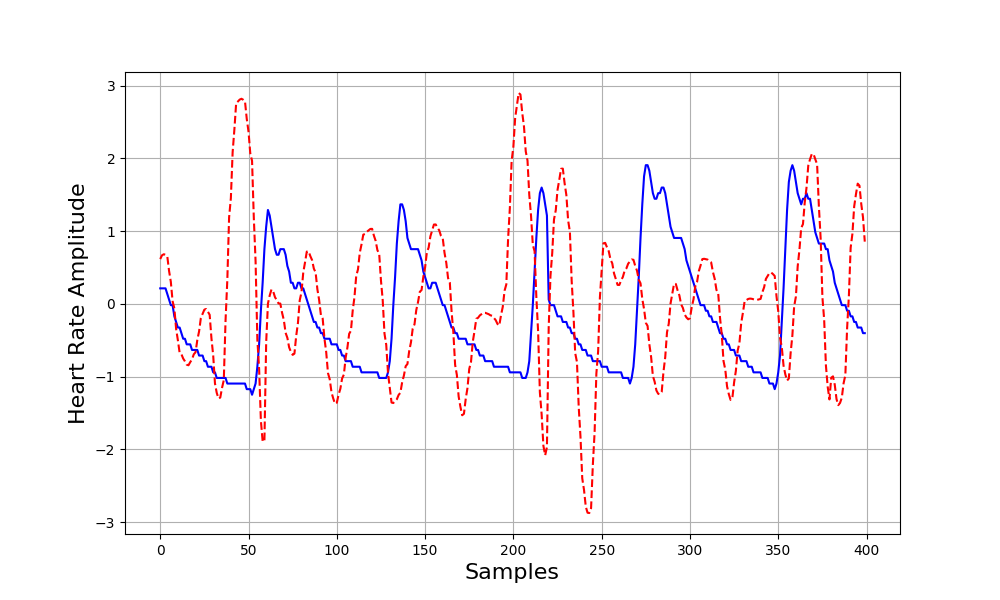}
}
\subfigure[TYrPPG]{
    \includegraphics[height = 35mm, width = 55mm]{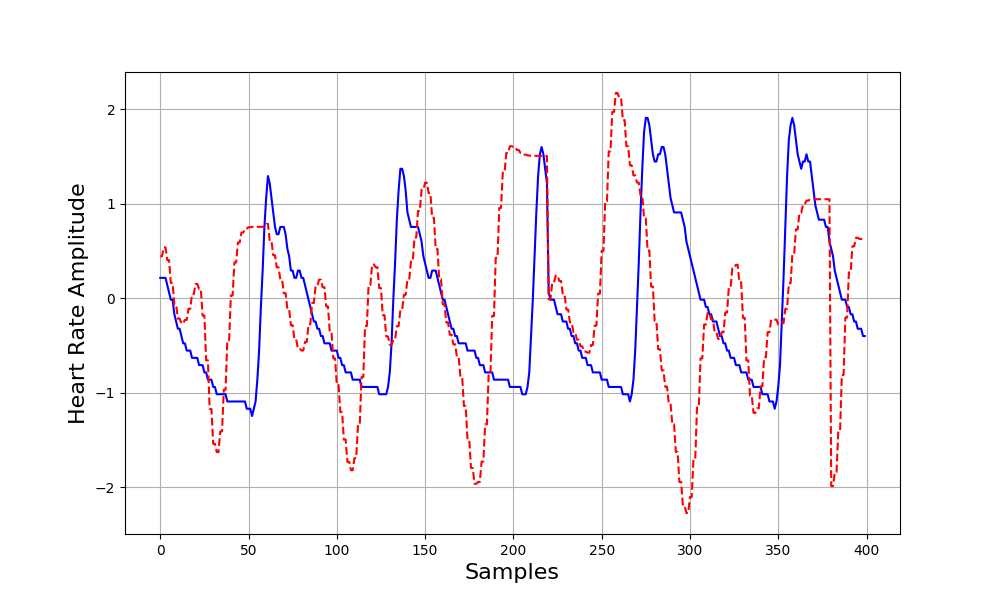}
}

    \caption{Visualization of the heart rate signals estimated by Physformer, RythmFormer, and TYrPPG on the PURE dataset. TYrPPG learns significantly better than the other two models. Straight and dotted lines mark the signals as ground truth and model estimation, respectively.}
    \label{subfig:vis}
\end{figure*}

In the cross-dataset experiment, there are two group experiments to evaluate the robustness and generalization abilities of the proposed model. According to Table \ref{tab:inter1}, it is obvious that TYrPPG performs best in the MMPD dataset after training on the PURE dataset compared with other baselines, which indicates its significant generalization ability.
 
\subsection{Visualization Analysis}
Based on Figure~\ref{subfig:vis}, we can discover that TYrPPG can significantly better learn the signal compared with the other two models. In particular, it can better capture the heart rate peak values from the video, showing its significant estimation ability. Since the peak heart rate has a higher diagnostic significance, TYrPPG provides more meaningful medical diagnosis results, showing its promising future.
\subsection{Ablation Study}
We have conducted an ablation study in the PURE dataset and set its train-test split ratio to 6:4 to evaluate the effectiveness of the proposed CSL and WSL. Based on the results of Table~\ref{tab:ablation}, the CSL has the best learning abilities compared with others, which indicates the necessity to combine the three loss items as the CSL. The results of WSL show the prospect that the model can learn the heart rate without knowing the details of the ground truth.
    
\section{Concluding Remarks}
We have proposed a novel rPPG for remotely estimating the heart rate. Compared with the existing method, our model structure is uncomplicated but powerful based on the proposed GVB, which combines the TSM and our modified Mambaout structure. Moreover, we have introduced a novel video MMD loss to enhance the learning of the ground truth distribution. The proposed CSL can provide both the details of the ground truth and the distribution similarity for the training model. TYrPPG achieves SOTA performance in two commonly used datasets, indicating its promising future. Further research may explore more about GVB block.

\bibliographystyle{ieeetr}
\bibliography{sample-base}

\end{document}